\documentclass[sigconf]{acmart}

\usepackage{graphicx}
\usepackage{subfiles}

\usepackage{todonotes}
\usepackage{booktabs}
\PassOptionsToPackage{hyphens}{url}\usepackage{hyperref}

\usepackage{dsfont}
\usepackage{amsmath}

\usepackage{caption}
\usepackage{subcaption}

\usepackage{multirow}
\usepackage{booktabs}
\usepackage{multicol}

\usepackage{soul}

\usepackage{xcolor}

\newcommand{\systemname}{MMMORRF}

\AtBeginDocument{%
  \providecommand\BibTeX{{%
    \normalfont B\kern-0.5em{\scshape i\kern-0.25em b}\kern-0.8em\TeX}}}

\copyrightyear{2025}
\acmYear{2025}
\setcopyright{rightsretained}
\acmConference[SIGIR '25]{Proceedings of the 48th International ACM SIGIR Conference on Research and Development in Information Retrieval}{July 13--18, 2025}{Padua, Italy}
\acmBooktitle{Proceedings of the 48th International ACM SIGIR Conference on Research and Development in Information Retrieval (SIGIR '25), July 13--18, 2025, Padua, Italy}\acmDOI{10.1145/3726302.3730157}
\acmISBN{979-8-4007-1592-1/2025/07}

\begin{document}



\title[MMMORRF: Multimodal Multilingual MOdularized Reciprocal Rank Fusion]{MMMORRF: \\ Multimodal Multilingual MOdularized Reciprocal Rank Fusion}




\settopmatter{authorsperrow=4}

\author{Saron Samuel}
\affiliation{%
  \institution{Stanford University}
  \city{Stanford}
  \state{CA}
  \country{USA}
}
\email{sdsam@stanford.edu}

\author{Dan DeGenaro}
\affiliation{%
  \institution{Georgetown University}
  \city{Washington, DC}
  \country{USA}
}
\email{drd92@georgetown.edu}

\author{Jimena Guallar-Blasco}
\affiliation{%
  \institution{Johns Hopkins University}
  \city{Baltimore}
  \state{MD}
  \country{USA}
}
\email{jgualla1@jhu.edu}

\author{Kate Sanders}
\affiliation{%
  \institution{Johns Hopkins University}
  \city{Baltimore}
  \state{MD}
  \country{USA}
}
\email{ksande25@jhu.edu}

\author{Oluwaseun Eisape}
\affiliation{%
  \institution{UC Berkeley}
  \city{Berkeley}
  \state{CA}
  \country{USA}
}
\email{eisape@berkeley.edu}

\author{Tanner Spendlove}
\affiliation{%
  \institution{BYU}
  \city{Provo}
  \state{UT}
  \country{USA}
}
\email{ths29@byu.edu}

\author{Arun Reddy}
\affiliation{%
  \institution{Johns Hopkins University}
  \city{Laurel}
  \state{MD}
  \country{USA}
}
\email{areddy24@jhu.edu}

\author{Alexander Martin}
\affiliation{%
  \institution{Johns Hopkins University}
  \city{Baltimore}
  \state{MD}
  \country{USA}
}
\email{amart233@jhu.edu}

\author{Andrew Yates}
\affiliation{%
  \institution{Johns Hopkins University}
  \city{Baltimore}
  \state{MD}
  \country{USA}
}
\email{andrew.yates@jhu.edu}

\author{Eugene Yang}
\affiliation{%
  \institution{Johns Hopkins University}
  \city{Baltimore}
  \state{MD}
  \country{USA}
}
\email{eugene.yang@jhu.edu}

\author{Cameron Carpenter}
\affiliation{%
  \institution{Johns Hopkins University}
  \city{Baltimore}
  \state{MD}
  \country{USA}
}
\email{ccarpe18@jhu.edu}

\author{David Etter}
\affiliation{%
  \institution{Johns Hopkins University}
  \city{Baltimore}
  \state{MD}
  \country{USA}
}
\email{detter2@jhu.edu}

\author{Efsun Kayi}
\affiliation{%
  \institution{Johns Hopkins University}
  \city{Laurel}
  \state{MD}
  \country{USA}
}
\email{ekay1@jhu.edu}

\author{Matthew Wiesner}
\affiliation{%
  \institution{Johns Hopkins University}
  \city{Baltimore}
  \state{MD}
  \country{USA}
}
\email{mwiesne2@jhu.edu}

\author{Kenton Murray}
\affiliation{%
  \institution{Johns Hopkins University}
  \city{Baltimore}
  \state{MD}
  \country{USA}
}
\email{kenton@jhu.edu}

\author{Reno Kriz}
\affiliation{%
  \institution{Johns Hopkins University}
  \city{Baltimore}
  \state{MD}
  \country{USA}
}
\email{rkriz1@jhu.edu}

\renewcommand{\shortauthors}{Samuel et al.}

\begin{abstract}

Videos inherently contain multiple modalities, including visual events, text overlays, sounds, and speech, all of which are important for retrieval. However, state-of-the-art multimodal language models like VAST and LanguageBind are built on vision-language models (VLMs), and thus overly prioritize visual signals. Retrieval benchmarks further reinforce this bias by focusing on visual queries and neglecting other modalities. 
We create a search system \systemname\ that extracts text and features from both visual and audio modalities and integrates them with a novel modality-aware weighted reciprocal rank fusion. 
\systemname\ is both effective and efficient, demonstrating practicality in searching videos based on users' information needs instead of visual descriptive queries. 
We evaluate \systemname\ on MultiVENT 2.0 and TVR, two multimodal benchmarks designed for more targeted information needs, and find that it improves nDCG@20 by 81\% over leading multimodal encoders and 37\% over single-modality retrieval.
\end{abstract}

\vspace{-1mm}




\maketitle

\section{Introduction} \label{sec:intro}

Online content has increasingly shifted toward video, many of which are multilingual and span multiple modalities, including visual events, spoken audio, embedded text, and non-speech sounds, such as music. As a result, users now require retrieval systems that handle both linguistic and modality diversity. Current commercial systems, such as YouTube search, rely heavily on non-visual metadata like titles, descriptions, and user engagement signals. While useful, these features are often incomplete or missing. To address this, we propose a multilingual video retrieval system that relies solely on visual, audio, and embedded textual content.

\begin{figure*}[t]
    \centering
    \includegraphics[width=\linewidth]{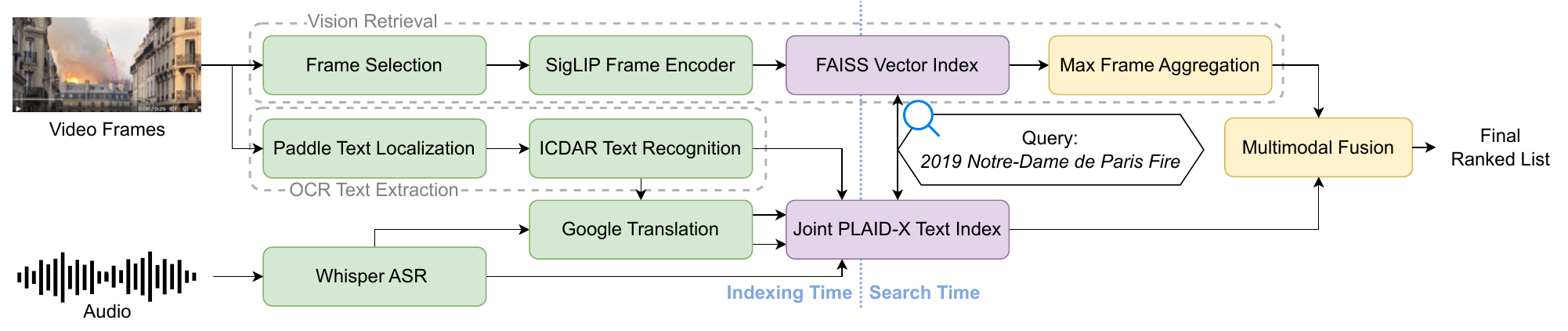}
    \caption{Diagram of \systemname\ pipeline and fusion system for event-centric video retrieval. Components to the left of the blue dotted line are processed at indexing time; components to the right are processed at search time.}
    \label{fig:pipeline}
\end{figure*}

Prior research has shown strong results using vision-language models~\cite{chen2023vast,zhu2024languagebindextendingvideolanguagepretraining,wang2024internvideo2} on academic benchmarks like \texttt{VALOR-32k}~\cite{chen2023valorvisionaudiolanguageomniperceptionpretraining} and \texttt{MSR-VTT}~\cite{xu2016msrvtt}. However, these datasets are small, English-centric, and rely on descriptive captions that do not reflect real-world search behaviors. Rather than relying on large vision and audio-language models, we introduce a practical and efficient fusion search engine, \systemname, that leverages information extracted using mature modality-specific technologies. Figure~\ref{fig:pipeline} illustrates our approach, which achieves state-of-the-art results on \texttt{MultiVENT 2.0}, with more than 80\% improvement over the best vision-language system.

Our contributions are three-fold: (1) we present, \systemname, a system that combines state-of-the-art components for multilingual, event-centric video retrieval; (2) we introduce a novel modality-aware fusion method for integrating multiple modalities; and (3) we provide a comprehensive comparison with vision-language models, demonstrating the effectiveness of our approach on a more realistic video retrieval task.
A video demonstrating \systemname\ is available on YouTube\footnote{\url{https://youtu.be/7jLisTeSjKM}}, and the implementation is available on GitHub. \footnote{\url{https://github.com/hltcoe/video-retrieval-demo}}
\section{Background}

\textbf{Multilingual and cross-language retrieval.}
Multilingual information retrieval (MLIR) involves retrieving documents in multiple languages using queries in a different language~\cite{oard1998survey, neuclir23}. Cross-language information retrieval (CLIR) is a specialized case where the corpus is monolingual but queries are in another language~\cite{neuclir22}. Recent advancements focus on end-to-end retrieval systems that eliminate translation dependencies~\cite{nair2022colbertx, li2022learning}, improving scalability in multilingual settings~\cite{lawrie2023mtt}. Techniques such as language model pretraining~\cite{yang2022c3}, data curation and translation~\cite{nair2022colbertx, nair2023blade}, and knowledge distillation~\cite{li2022learning, yang2024td, yang2024mtd} have enhanced the effectiveness of neural retrieval models~\cite{neuclir23, yang2024mtd}.

\textbf{Cross-modal retrieval.}
Cross-modal retrieval retrieves data in one modality using queries from another, such as retrieving videos using textual queries. Traditional methods treat each modality independently~\cite{wang2016comprehensive}, while modern deep learning enables joint multimodal representations, allowing direct comparisons across modalities. Models based on Contrastive Language–Image Pretraining (\texttt{CLIP})~\cite{radford2021learningtransferablevisualmodels} have advanced this field and are further adapted for video retrieval~\cite{fang2021clip2videomasteringvideotextretrieval, jin2022expectationmaximization, xue2023clipvipadaptingpretrainedimagetext}. However, most existing retrieval models remain limited to single-language and single-modality settings, whereas real-world online content spans multiple languages and integrates textual, auditory, and visual information, posing new challenges for retrieval systems.

\textbf{Video retrieval.}
Video retrieval involves identifying the most relevant videos for a given query~\cite{cao2024rapefficienttextvideoretrieval, tang2024musemambaefficientmultiscale}. Early systems relied on metadata such as titles and descriptions~\cite{wactlar2002digital}, but deep learning has enabled retrieval using visual, audio, and embedded text features~\cite{bain2021frozen, fang2021clip2video}. Recent multimodal systems integrate these signals via fusion transformers~\cite{chen2023valorvisionaudiolanguageomniperceptionpretraining, wang2024internvideo2}. However, most existing benchmarks focus on monolingual collections and simple queries~\cite{sanders2024survey}, making the task of true multilingual, multimodal retrieval particularly challenging to evaluate. Extracted video text has been explored in monolingual retrieval~\cite{wu2025large}, video question-answering~\cite{zhao2022towards}, and captioning~\cite{wu2023cap4video}, but robust multilingual text extraction for video retrieval remains underexplored~\cite{sanders2023multiventmultilingualvideosevents}.

\section{Video Retrieval Pipeline}

In this section, we present the components of our \systemname\ retrieval system for multilingual video retrieval. 

\noindent
\textbf{Visual Content.} 
To capture signals from the visual content, we develop a sub-pipeline that selects, encodes, and searches video frames, illustrated at the top of Figure~\ref{fig:pipeline}. To balance efficiency and effectiveness, we uniformly sample 16 frames per video. 
Through pilot testing on MultiVent 2.0 training set~\cite{MultiVENT-2.0}, we found that the benefit of sampling more frames saturates dramatically after 16. 
We also investigated key-frame detection tools, such as PySceneDetect~\cite{PySceneDetect}, but found no substantial difference between uniform sampling. 

To bridge the modality gap between textual queries and video frames, we use \texttt{SigLIP}~\cite{zhai2023sigmoidlosslanguageimage}, a variant of \texttt{CLIP} that replaces the standard contrastive learning softmax normalization with a sigmoid loss function. For efficient retrieval, frame embeddings are indexed using \texttt{FAISS}, a scalable vector search engine~\cite{douze2024faiss}. At search time, queries are encoded using \texttt{SigLIP}’s text encoder and matched against the indexed frame embeddings by computing the maximum query-frame score (\texttt{MaxFrame}) for each video/query pair.

\noindent
\textbf{Video Optical Character Recognition (OCR).} 
To accurately search embedded text in videos, such as overlay text from news broadcasts or street signs, we utilize an OCR sub-pipeline consisting of three steps: frame selection, text localization (or detection) and text recognition. For frame selection, we again select 16 uniformly sampled frames. Text localization is performed using the open-source \texttt{PaddleOCR} toolkit~\cite{liao2020real,liao2022real}. The localized line images are then cropped and recognized using the multilingual OCR system described in \citet{icdar_ocr}, which combines a vision transformer encoder trained with a Connectionist Temporal Classification (\texttt{CTC}) objective and an auto-regressive, character-level decoder.

\noindent
\textbf{Audio Transcription.} 
We use OpenAI’s \texttt{Whisper Large-v2}, a state-of-the-art multilingual automatic speech recognition (ASR) model, to transcribe video audio~\cite{pmlr-v202-radford23a}. \texttt{Whisper} is a transformer-based sequence-to-sequence model pre-trained on 680,000 hours of multilingual audio for several general speech tasks, including language identification, time-aligned transcription, and time-aligned any-to-English translation. Its multilingual capabilities and ability to handle non-speech audio make it particularly well-suited for the diverse audio content found in video collections.

\noindent
\textbf{Extracted text Retrieval.} 
To search the text extracted from visual and audio content, we employ \texttt{multilingual ColBERT-X}~\cite{nair2022colbertx, lawrie2023mtt}, a multilingual cross-language variant of \texttt{ColBERT} with \texttt{PLAID-X}~\cite{yang2024td}\footnote{\url{https://github.com/hltcoe/ColBERT-X}} implementation. During indexing, each piece of extracted text, i.e., ASR and OCR outputs, is encoded into multiple vectors and preprocessed with \texttt{K}-means clustering to enable fast approximate search. At search time, \texttt{PLAID-X} encodes the queries and performs contextual late interaction between the query and text representations using product quantization. Both the extracted text and queries are encoded using a multilingual \texttt{ColBERT-X}~\cite{yang2024mtd} model to ensure robust processing across languages.\footnote{\url{https://huggingface.co/hltcoe/plaidx-large-eng-tdist-mt5xxl-engeng}} As shown in Figure~\ref{fig:pipeline}, the OCR and ASR outputs are jointly fed into \texttt{ColBERT-X} at index time, referred to as \texttt{Joint Index} (\texttt{JI}) in Table~\ref{tab:main_results}.

\noindent
\textbf{Multimodal Retrieval Late Fusion.}
To leverage the strengths of both text-based and vision-based retrieval methods, we implement a modality-aware Weighted Reciprocal Rank Fusion (\texttt{WRRF}) approach. This fusion method combines the rankings from the text retrieval system (\texttt{PLAID-X}) and the vision retrieval system (\texttt{SigLIP}), assigning importance weights based on each video’s characteristics. \texttt{WRRF} extends traditional Reciprocal Rank Fusion (\texttt{RRF})~\cite{cormack_rrf_2009}. 
Instead of summing the reciprocal ranks from both systems, \texttt{WRRF} computes a \textit{video-dependent} linear combination of the two ranks to produce the final ranking score, defined as follows:
\begin{align*}
    \texttt{WRRF}(q, d) = \frac{\alpha_d}{r_\text{text}(q, d) + k} + \frac{1 - \alpha_d}{r_\text{vision}(q, d) + k}
\end{align*}
where $\alpha_d$ is the coefficient for the combination. 
Given the query $q$ and video $d$, the ranks from the two systems are denoted as $r_{text}(q, d)$ and $r_{vision}(q, d)$, respectively, with a smoothing hyperparameter $k$. In our implementation, we set $k=0$, diverging from the typical value of $60$, to emphasize highly ranked results from either modality. This adaptive weighting strategy enables our system to capitalize on the strengths of each modality, enhancing retrieval performance over a diverse video collection. 

We estimate the weighting for each video based on the retrieval score of a predefined query using the \texttt{SigLIP} model, which reflects the relative reliability of text-based and vision-based retrieval. We operationalize the weightings based on how closely a video resembles professional news content, hypothesizing that news-like videos yield more informative ASR and OCR text.
Specifically, we employ the text query, \textit{news anchor live coverage broadcast microphone breaking news}, and follow our video retrieval pipeline, which retrieves at the frame level and aggregates results using \texttt{MaxFrame}. These scores are independent of any specific query and simply assess the characteristics of the videos, which means scoring occurs during the indexing phase. We argue that this preprocessing step, while it may appear ad hoc, is analogous to tokenization decisions in text retrieval pipelines. Different retrieval collections, particularly those from varying genres, necessitate distinct tokenization and text-cleaning processes to achieve optimal retrieval results.

\section{Experiment Setup}

\begin{table*}[t]
\centering

\newcommand{\mt}{MT>>}
\newcommand{\ti}{$\times$}
\setstcolor{red}
\newcommand{\x}[1]{\textsuperscript{\st{#1}}}

\caption{Retrieval effectiveness on \texttt{MultiVENT 2.0} and \texttt{TVR}. 
``JI'' in the fusion method column indicates joint index. 
Only values in \texttt{MultiVENT 2.0} are tested for statistical significance, employing a paired t-test with 95\% confidence under Bonferroni correction of 110 tests (all pairs of 11 systems and 2 metrics). 
Almost all differences are significant, with exceptions marked in superscript using a red strike line. 
}\label{tab:main_results}

\setlength\tabcolsep{0.5em}

\begin{tabular}{@{}ccc|cc|c|ll|ll@{}}
\toprule
{}  & \multicolumn{2}{c|}{Vision} & \multicolumn{2}{c|}{Audio} & Fusion
    & \multicolumn{2}{c|}{MultiVENT 2.0} & \multicolumn{2}{c}{TVR} \\
{}  & Frames &        OCR & Acoustic &        ASR & Method &  nDCG@10      &  R@10        &    R@1  &   R@10  \\
\midrule
a.  & InternVideo2 &  --- & ---      &   ---       & Embed. &  0.005        & 0.004        &    0.053  &  0.158    \\
b.  &   VAST &   ---      &     VAST &    ---      & Embed. &  0.035        & 0.042        &  0.100  &  0.268  \\
c.  &LangBind&       ---  &      --- &    ---      & Embed. &  0.324\x{g}   & 0.355\x{eg}  &  0.087  &  0.258  \\
\midrule
d.  & SigLIP &            &          &             &   ---  &  0.375\x{eg}  & 0.409\x{fh}  &  0.156  &  0.375  \\
e.  &        &       BM25 &          &             &   ---  &  0.370\x{d}   & 0.367\x{cg}  &  0.005  &  0.020  \\
f.  &        &    PLAID-X &          &             &   ---  &  0.415\x{h}   & 0.414\x{dh}  &  0.004  &  0.015  \\
g.  &        &            &          &        BM25 &   ---  &  0.347\x{cd}  & 0.313\x{ce}  &  0.200  &  0.381  \\
h.  &        &            &          &     PLAID-X &   ---  &  0.427\x{f}   & 0.425\x{df}  &  0.192  &  0.400  \\
\midrule
i.  &        &    PLAID-X &          &     PLAID-X &     JI &  0.551\x{j}   & 0.556        &  0.192 &  0.400  \\
j.  & SigLIP &    PLAID-X &          &     PLAID-X & JI+RRF &  0.562\x{i}   & 0.600        &  \textbf{0.232} &  0.537  \\
k.  & SigLIP &    PLAID-X &          &     PLAID-X &JI+WRRF & \textbf{0.586}&\textbf{0.611}&  0.201 &  \textbf{0.540} \\
\bottomrule
\end{tabular}

\end{table*}

We evaluate our proposed video retrieval pipeline on two datasets: \texttt{MultiVENT 2.0} and \texttt{TVR}. These datasets differ significantly in retrieval scenarios, providing a broad evaluation of our approach.

\textbf{\texttt{MultiVENT 2.0}}~\cite{MultiVENT-2.0} is a large-scale multilingual event-centric video retrieval dataset consisting of 218,000 videos and 3,906 manually curated queries covering diverse world events. The dataset is split into \texttt{MultiVENT-Train} (108,600 videos, 1,361 queries) and \texttt{MultiVENT-Test} (109,800 videos, 2,545 queries). Videos span professionally edited news reports to raw mobile footage, covering six primary languages: Arabic, Chinese, English, Korean, Russian, and Spanish. Unlike conventional video retrieval benchmarks, which often use captions as queries, \texttt{MultiVENT 2.0} emphasizes real-world information-seeking behavior, making it more analogous to ad hoc text retrieval tasks like \texttt{MS MARCO}~\cite{bajaj2018msmarco}.

\textbf{\texttt{TVR}} (TV show Retrieval) is a multimodal video retrieval dataset consisting of 109K queries on 21.8K videos from six TV shows ~\cite{lei2020tvr}. The dataset requires systems to understand both videos and subtitles. The dataset is split into 80\% train, 10\% val, 5\% test-public and 5\% test-private splits. We use the validation split in our analysis. Each query is written using either the subtitles, video, or both, and is linked to a specific temporal window, requiring precise moment localization within videos.

All extracted text from \texttt{MultiVENT 2.0} is concatenated with their machine translation obtained through Google Translate before PLAID-X indexing, which is a common practice in CLIR~\cite{neuclir22, neuclir23}. For \texttt{TVR}, since the videos are already in English, we do not obtain additional text before indexing. Following common video retrieval evaluation, we report Recall at 1 and 10 (\texttt{R@1} and \texttt{R@10}) for both datasets, as well as normalized Discounted Cumulative Gain (\texttt{nDCG@10}) for \texttt{MultiVENT 2.0} to assess ranked retrieval performance.

In addition to pipeline approaches, we report results using three end-to-end multimodal retrieval models that leverage pretrained vision-language models (VLMs): \texttt{InternVideo2}~\cite{wang2024internvideo2}, \texttt{VAST}~\cite{chen2023vastvisionaudiosubtitletextomnimodalityfoundation}, and \texttt{Language-Bind}~\cite{zhu2024languagebindextendingvideolanguagepretraining}. 
These models have recently achieved state-of-the-art results on various vision tasks such as visual question answering and text-video retrieval, though often on small, unrealistic datasets. Most VLMs feature separate encoders for different modalities, while others attempt to use a unified encoder across modalities.

We use eight NVIDIA V100 GPUs for \texttt{PLAID-X} indexing and \texttt{SigLIP} encoding for faster parallel encoding. We ran \texttt{FAISS} without any compression and approximation. 
Serving queries can be done without a GPU for PLAID-X, but it is necessary for SigLIP.
Therefore, at query serving time, we also use a V100 GPU for encoding the queries. 
\section{Results and Analysis}

Table~\ref{tab:main_results} summarizes the evaluation results of both the baselines and variants of \systemname\ on \texttt{MultiVENT 2.0} and \texttt{TVR}. While \texttt{InternVideo2}, \texttt{VAST}, \texttt{LanguageBind} are strong vision retrieval models with some decent effectiveness on \texttt{TVR}, their performance on \texttt{MultiVENT 2.0} is notably poor. \texttt{InternVideo2}'s results are particularly surprising, given that it was trained on much of the \texttt{MultiVENT 2.0} collection. This underperformance likely stems from its reliance on automatically generated captions, reinforcing our argument that \texttt{MultiVENT 2.0} presents a more realistic and challenging video retrieval task.

Pipelines that independently pass OCR and ASR outputs through mature text retrieval systems yield \texttt{nDCG@10} scores of 0.415 and 0.427 on \texttt{MultiVENT 2.0}, respectively, both outperforming the two vision-language models. 
In \texttt{TVR}, where the videos are already in English, searching with extracted text still outperforms \texttt{VAST}. 
Here, PLAID-X does not provide much improvement, indicating that serving these queries can merely rely on lexical matches. 

Jointly indexing both OCR and ASR outputs achieves an \texttt{nDCG@10} of 0.551, outperforming all non-translated single-modality systems. 
Interestingly, combining the text sub-pipeline with the visual retrieval system (\texttt{SigLIP}) using ordinary reciprocal rank fusion (\texttt{RRF}) yields only a marginal improvement without statistical significance. However, applying the video-dependent weighted \texttt{RRF} (\texttt{WRRF}) we developed yields a statistically significant 4.2\% improvement (from 0.562 to 0.586 \texttt{nDCG@10}). Notably, incorporating \texttt{SigLIP} into the pipeline results in a statistically significant 6.4\% improvement (from 0.551 to 0.586) while introducing minimal additional latency, as discussed in Section~\ref{sec:results:timing}.

While end-to-end models have been the focus of video retrieval research, our results demonstrate that a pipeline approach composed of mature components remains both more effective and practical for real-world applications—especially when evaluated on a large-scale, realistic benchmark like \texttt{MultiVENT 2.0}.

\subsection{Encoding Throughput and Query Latency}\label{sec:results:timing}

\begin{table}[t]
\caption{Preprocessing time (per video) and query latency (per query). The overall timing assumes parallelism when possible. 
}\label{tab:timing}

\centering
\setlength\tabcolsep{0.27em}
\begin{tabular}{r|c|cccc|c}
\toprule
                    &   Vision &    OCR &   ASR &            MT &  PLAID-X &   Overall \\
\midrule
Preprocess / video  &       1s &    30s &   36s &        < 0.5s &   < 0.5s &       37s \\
Search / query      &    200ms &    --  &   --  &          --   &    300ms &     300ms \\
\bottomrule
\end{tabular}
\vspace{-1em}
\end{table}

We report both preprocessing (encoding and indexing) and search times in Table~\ref{tab:timing}. Since the OCR and ASR processes can be run in parallel, the overall text retrieval pipeline takes approximately 37 seconds per video. \texttt{SigLIP} video frame encoding and indexing, the other major component, requires only 1 second per video and runs concurrently with the text pipeline. At search time, the combination of the \texttt{FAISS} index and \texttt{PLAID-X} retrieval system results in an overall per-query latency of approximately 300 ms, as both the vision and text pipelines operate in parallel. This demonstrates that our system is both effective and efficient. A query latency of 300 ms is unlikely to be noticeable to users, making our system highly practical for real-world search applications~\cite{arapakis2014impact}.

\section{Conclusion}

In this work, we present \systemname\ -- a pipeline and fusion system for event-centric multimodal retrieval, which we argue is a more practical framing for video retrieval tasks. Unlike prior approaches that focus primarily on descriptive visual information, our system extracts specific text from videos via Automatic Speech Recognition and Optical Character Recognition and processes this information with a mature multilingual neural retrieval system. Evaluating on \texttt{MultiVENT 2.0}, a large-scale multilingual collection, we demonstrate that our pipeline and fusion system achieves a new state-of-the-art, outperforming the best vision-language model by 81\%, and the top-performing single-modality system by 37\%. Notably, our approach also achieves a query serving time of under 300 ms, making it both high-performing and efficient. To the best of our knowledge, this is the first scholarly work to integrate modern neural video encoding and neural retrieval models into a practical and deployable system for ad hoc video retrieval. We believe our approach lays the groundwork for building scalable, real-time video retrieval systems capable of addressing the complexities of multilingual, multimodal event-centric search tasks.


\bibliographystyle{ACM-Reference-Format}
\bibliography{bibio}


\end{document}